# A PICTURE IS WORTH A COLLABORATION: ACCUMULATING DESIGN KNOWLEDGE FOR COMPUTER-VISION-BASED HYBRID INTELLIGENCE SYSTEMS

*Research Paper*

Zschech, Patrick, Friedrich-Alexander-Universität Erlangen-Nürnberg, Germany, patrick.zschech@fau.de

Walk, Jannis, Karlsruhe Institute of Technology, Karlsruhe, Germany & Ceratizit Austria GmbH, Reutte, Austria, jannis.walk@kit.edu

Heinrich, Kai, Otto-von-Guericke-Universität Magdeburg, Germany, kai.heinrich@ovgu.de

Vössing, Michael, Karlsruhe Institute of Technology, Germany, michael.voessing@kit.edu

Kühl, Niklas, Karlsruhe Institute of Technology, Karlsruhe, Germany & IBM, Ehningen, Germany, niklas.kuehl@kit.edu

## Abstract

*Computer vision (CV) techniques try to mimic human capabilities of visual perception to support labor-intensive and time-consuming tasks like the recognition and localization of critical objects. Nowadays, CV increasingly relies on artificial intelligence (AI) to automatically extract useful information from images that can be utilized for decision support and business process automation. However, the focus of extant research is often exclusively on technical aspects when designing AI-based CV systems while neglecting socio-technical facets, such as trust, control, and autonomy. For this purpose, we consider the design of such systems from a hybrid intelligence (HI) perspective and aim to derive prescriptive design knowledge for CV-based HI systems. We apply a reflective, practice-inspired design science approach and accumulate design knowledge from six comprehensive CV projects. As a result, we identify four design-related mechanisms (i.e., automation, signaling, modification, and collaboration) that inform our derived meta-requirements and design principles. This can serve as a basis for further socio-technical research on CV-based HI systems.*

## 1 Introduction

Intelligent information systems (IS) that incorporate artificial intelligence (AI), so-called AI-based IS (Maedche et al. 2019), have a massive impact on our society and revolutionize how we live, act, and work together. Cars begin to drive autonomously in real traffic (Grigorescu et al. 2020); smart home systems recognize and adapt to individual user preferences (Fischer et al. 2020); and medical assistance systems support doctors in diagnosing hard-to-find diseases (McKinney et al. 2020). A key enabler for the realization of such scenarios is the capability of modern AI-based systems to automatically process high-dimensional data to identify useful patterns and relationships that can be utilized for decision support and business automation purposes (Brynjolfsson and Mcafee 2017).

An important sub-field in this context is the area of computer vision (CV). It seeks to automatically extract useful information from images to mimic human capabilities of visual perception (Szeliski 2010). On this basis, time-consuming and labor-intensive tasks like the recognition, detection, localization, tracking, and counting of objects can be supported more efficiently to save unnecessary resources and relieve the burden of human workers (Heinrich, Roth, et al. 2019).

Fueled by the broad availability of huge online image databases and the broad access to necessary computing power, the field of CV is currently experiencing a considerable phase of scientific progress and dissemination expressed by manifold activities in research and practice. As such, we can observe a continuous development of advanced algorithms based on machine learning and especially artificial neural networks (ANN) (Bharati and Pramanik 2020); procedure models and tutorials for solution development are proposed (Griebel, Dürr, et al. 2019); and a global community of developers shares reusable software code and provides user-friendly programming frameworks (Chollet 2017). As a result, more and more CV systems are being embedded into organizational and societal contexts across a wide range of domains, such as traffic surveillance (Liu et al. 2017), manufacturing (Wang et al. 2018), agriculture (Tian et al. 2020), and sports (Thomas et al. 2017).

However, past efforts in research and practice often exclusively focused on technical performance aspects when designing and developing CV systems (e.g., achieved accuracy, required computing resources), while neglecting socio-technical facets, such as transparency, control, and autonomy. Coming from an IS research perspective, such aspects are crucial, for example, to ensure that a technology is accepted by its users and that it is in line with the organization's objectives (Schaper and Pervan 2005). Especially in the realm of designing and working with AI systems, such aspects play a fundamental role and therefore should be translated into an AI-based system design to aid the user in setting up, understanding and using autonomously operating AI systems (Thiebes et al. 2020). Dellermann, Ebel, et al. (2019) discuss the concept of hybrid intelligence (HI) that combines the complementary strengths of both sides in order to reach superior performance than would be achievable separately. Although this hybrid system design can bring out the best of both worlds, it is faced with challenges like algorithm aversion that occur due to the complexity of the AI system resulting in the distinctness of human and AI system in such a setting (Berger et al. 2021).

Against this background, this paper deals with the design of computer-vision-based hybrid intelligence systems (CV-HIS). The aim is to derive prescriptive design knowledge from a socio-technical view that should ultimately result in a (nascent) design theory (Gregor and Jones 2007). To this end, we follow a design science research (DSR) approach (Hevner et al. 2004) and reflect on accumulated design knowledge generated in six comprehensive CV development projects. With this strategy, we pursue a reflective approach (Möller et al. 2020) based on real problem cases as encountered and solved in practical settings (Iivari 2015). More specifically, we contribute in the following ways: (i) we conceptualize the HI collaboration in the realm of vision-based tasks to identify design-related mechanisms, (ii) we derive requirements and abstract them to meta-requirements in relation to central kernel theories in terms of justificatory knowledge, and (iii) we accumulate design principles by abstracting from specific design features which were implemented across the different CV projects.

Our paper is structured accordingly: In Section 2, we introduce the foundations and refer to related work. Subsequently, we depict our research approach in more detail in Section 3. In Section 4, we describe our selected CV cases, followed by the conceptualization of CV-HIS in Section 5. We then proceed in Section 6 to outline our derived design knowledge by distinguishing between four identified mechanisms as introduced by design. Finally, we summarize and discuss our contribution and present an outlook of further research opportunities in Section 7.

## 2 Foundations and Related Work

### 2.1 Computer Vision and Artificial Intelligence

The field of CV is concerned with the development of techniques for the acquisition, processing, analysis, and understanding of digital images to transform high-dimensional data into symbolic or numerical information (e.g., for automated decision-making). Just as humans use their eyes and brains to understand the world around them, CV attempts to produce the same effect so that computers can perceive and understand an image or a sequence of images and act accordingly in each situation. This understanding can be achieved by disentangling high-dimensional data from images using models built with the aid of geometry, statistics, physics, and learning theory (Forsyth and Ponce 2002). Driven by personal or industrial motives, grand advances have been made in several areas such as optical character recognition, machine inspection, 3D model building, disease diagnostics, motion capture, or surveillance (Szeliski 2010). Nowadays, CV tasks are increasingly performed by AI-based systems that rely on machine learning algorithms. Of particular interest are ANNs, which can be organized in deep network architectures consisting of multiple, hierarchical processing layers (Janiesch et al. 2021). This allows them to automatically process spatial information in raw image data and learn patterns that are relevant for prediction tasks, which is often also referred to as deep learning (DL) (LeCun et al. 2015). The automated learning of patterns by DL models is usually done in a supervised manner. This means that humans provide training data that are tagged with labels/annotations to specify the target of the learning task (Hastie et al. 2009; Sager et al. 2021). Computer vision is also gaining momentum in the field of IS. Extant publications range from use cases like road crack detection (Chatterjee et al. 2018) and automated fashion recommendations (Griebel, Welsch, et al. 2019) to guidelines for the technical aspects of CV projects (Griebel, Dürr, et al. 2019).

## 2.2 Hybrid Intelligence

Due to the rising capabilities of AI in the last decade (Russell 2016), researchers are increasingly reconsidering much of the established design knowledge regarding how intelligent systems should be designed. As noted by Zheng et al. (2017), the development of AI is profoundly changing how humans interact with their environment and how they support their work processes. The authors introduce *hybrid-augmented intelligence* as a means to combine "human cognitive ability and the capabilities of computers". Similarly, Dellermann, Ebel, et al. (2019) define *hybrid intelligence* as "the ability to achieve complex goals by combining human and artificial intelligence, thereby reaching superior results to those each of them could have accomplished separately". The core concept of augmenting both sides, computers and humans, and thus creating a symbiotic relationship between them is also mentioned by Akata et al. (2020), Maedche et al. (2019), and Seeber et al. (2020).

While the benefits of the relationships are depicted in the recent literature, it is also mentioned that to achieve beneficial utility; it is required to "[develop] novel interaction paradigms that exploit the strengths and overcome the weaknesses of both partners" (Terveen 1995). However, so far, only a few researchers have formalized the required design knowledge.

| **Reference** | **Approach** | **Focus** | **Application area** | **Derived design knowledge** |
|---|---|---|---|---|
| Dellermann, Ebel, et al. (2019) | Definitional | Conceptualization | General | n/a |
| Wang et al. (2019) | Behavioral | Understanding of the field | AutoAI | n/a |
| Chakraborti and Kambhampati (2018) | Behavioral | Mental models | Urban search and rescue | n/a |
| Zheng et al. (2017) | Behavioral | Conceptualization | Multiple | n/a |
| Dellermann, Lipusch, et al. (2019) | DSR Strategy I | Prescriptive knowledge | Business model validation | Requirements and design principles |
| This work | DSR Strategy II | Conceptualization and prescriptive knowledge | Computer vision | Meta-requirements and design principles |

*Table 1. Overview of related literature and positioning of this work.*

Zheng et al. (2017) provide a framework that suggests a human-in-the-loop approach comprising a computer with AI capabilities that outputs a prediction along with confidence scores as an uncertainty assessment and a human decision-maker. Additionally, it is suggested that the human gives feedback to the AI system through additional data labeling. In comparison to the rather human-centric approach, Schwartz et al. (2016) provide the concept of an augmented human in the context of industrial cyber-physical systems. The human is equipped with sensors and augment VR tools like gloves and glasses to be able to act on the AI system's suggestions and provide additional feedback through collected sensor data. Dellermann, Ebel, et al. (2019) provide a structured overview of design knowledge for HI systems in the form of a taxonomy, including the dimensions task characteristics as well as human and machine learning paradigms. As a specific example, Dellermann, Lipusch, et al. (2019) outline design principles for business model validation by suggesting to combine crowd-sourced data and machine predictions. Table 1 summarizes our findings of the related work of HI systems.

## 3      Research Approach

DSR is a fundamental paradigm in IS research concerned with the construction of socio-technical artifacts to solve organizational problems and derive generalizable design knowledge (Gregor and Hevner 2013). One of the ultimate goals of DSR is the formulation and consolidation of design theories (Beck et al. 2013). To communicate intermediate theoretical results of the theorizing process, so-called nascent design theories, Gregor and Jones (2007) describe important components, ranging from the purpose and scope (meta-requirements) and justificatory knowledge (kernel theories) to principles of form and function (design principles) and expository instantiations (implemented systems).

In order to derive generalizable design knowledge, two strategies are applicable (Iivari 2015). The first strategy ("Strategy I") deals with the construction of IT meta-artifacts as a generic solution concept for a problem class in advance. In the second strategy ("Strategy II"), abstract knowledge is derived in a reflective manner, i.e., specific IT artifacts are first designed and implemented within a practical context. Subsequently, generalizable knowledge emerges during or after the design iterations of the artifact when abstracting from the specific implementation (Iivari 2015; Möller et al. 2020). For the work at hand, we chose a reflective, practice-inspired approach in-line with Strategy II. More specifically, we analyzed CV development projects from industry and accumulated design knowledge from specific implementations while informing our findings with justificatory knowledge from well-established kernel theories to support the observed phenomena. In this way, it was intended to unveil implicit design knowledge concerning socio-technical mechanisms with respect to hybrid intelligence interactions, which so far have been little addressed in the literature on the design of CV systems.

To conduct our research, we organized a focus group with experts in the field (Morgan 1996). For this purpose, we identified researchers in the IS community fulfilling the following four criteria: (i) active involvement in multiple CV system development projects, (ii) fundamental understanding of AI-based technologies, (iii) sufficient experience in conducting DSR projects, and (iv) solid understanding of IS-related theories. As a result, we recruited six researchers from six different institutions which were asked to participate in a series of workshops to reflect their collected design knowledge during their involvement in CV-related development projects.

Due to COVID-19, the workshops were conducted using video conferencing tools. Initially, there was no predefined structure for the full series of workshops as it only became apparent during the sessions which consecutive steps were necessary to derive generalizable design knowledge. Overall, this resulted in a total of seven workshop sessions over a period of three months, each lasting between 60 and 120 minutes. To mitigate the bias caused by opinion leaders within the group, there were several tasks to be performed by each researcher individually before presenting and reflecting the results in each workshop with the entirety of the group. The full series of workshops is summarized in Figure 1 and is briefly described below with regard to the performed activities and achieved results.

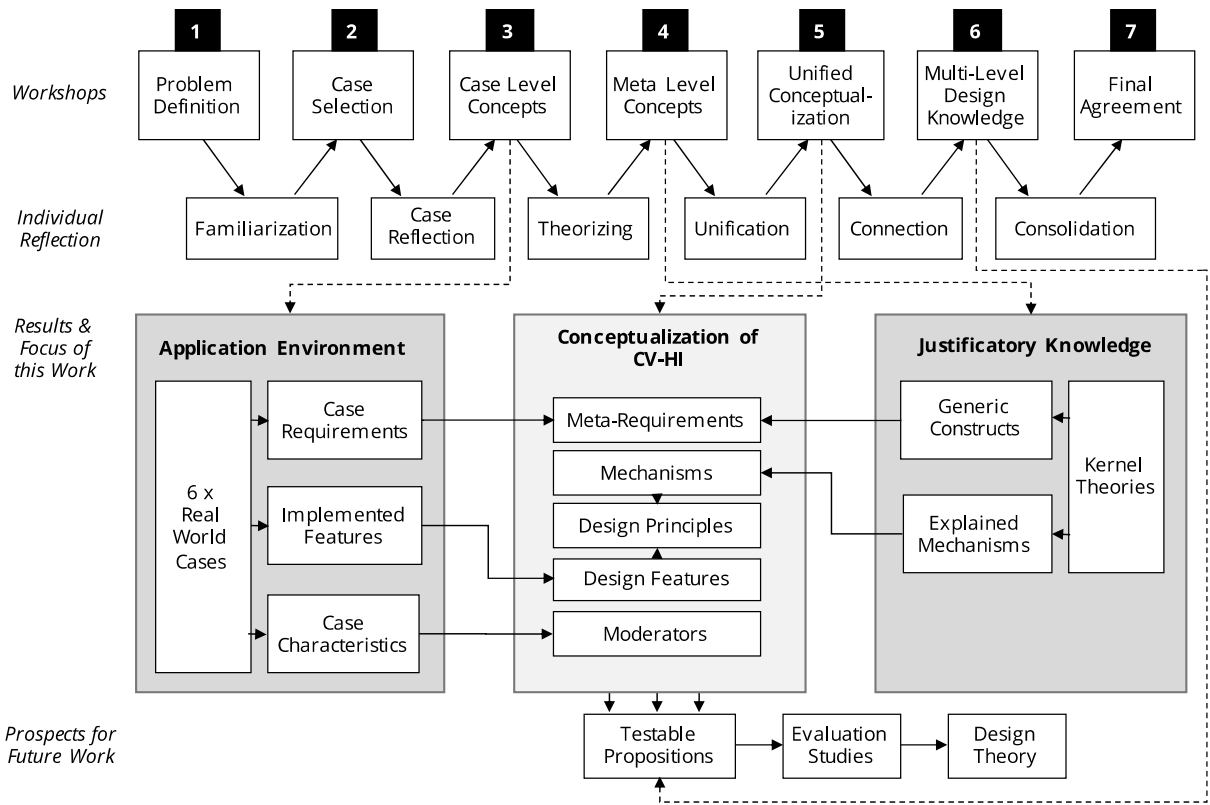

*Figure 1.     Applied research approach and focus of this paper.*

In the **first session**, the project's initiator introduced the research idea and the group agreed upon a common understanding of relevant fundamentals (e.g., CV/AI/HI technology, understanding of generalizable design knowledge) to establish a shared problem definition. On this basis, all participants had to prepare initial proposals on how design knowledge could be derived and presented systematically by using their individual experiences from research and practice.

In the **second session**, the individual proposals were discussed, which finally resulted in the decision to pursue a reflective methodological strategy by starting from a selection of practical cases and then incrementally accumulating and generalizing design knowledge towards the development of a nascent design theory. Thus, a total of six CV development cases were chosen as a basis for knowledge generation, which subsequently had to be reflected and characterized by the individual researchers.

In the **third session**, the results were presented to the group to derive a collection of (i) case requirements, (ii) implemented features, and (iii) several more case characteristics (e.g., domain, criticality, expertise of users) to classify each specific case (cf. Section 4). By comparing similarities and differences across all projects, various case level concepts could be extracted and tentatively connected to gain a first level of abstraction towards generalizable design knowledge. This included known concepts like supervised learning, active learning, degree of automation, accuracy, speed, image complexity, labeling cost, black-box behavior, explanation, uncertainty, compliance, and others. On this basis, all researchers were asked to conduct an individual step of theorizing. To this end, each participant should prepare conceptualizations and schemata to link theoretical constructs and mechanisms with observations collected within the cases.

The results were jointly discussed in the **fourth session**. Several potential kernel theories were identified, such as technology acceptance theory (Venkatesh et al. 2012), principal-agent and signaling theory (Eisenhardt 1989), and algorithm aversion (Dietvorst et al. 2015). In combination, they resulted in several meta-level concepts like performance, effort, system restrictiveness, information asymmetry, perceived control, trust, and others. However, the participants struggled to connect observations from the application environment with justificatory knowledge from theory. This led to a divergent picture within the individual conceptualizations, which were created with different emphases on different levels of abstraction. For example, some participants focused on HI interaction patterns between the human and the computer from different theoretical lenses, while others connected specific case properties, features, and requirements to theoretical constructs. As a result, three researchers agreed to translate the different views into a unified conceptualization to serve as a foundation for further discussion.

In the **fifth session**, the unified conceptualization was discussed and incrementally refined by the entire group. With this unified framing, three distinct design-related components could be distinguished from each other towards an adequate level of abstraction of derived design knowledge, precisely (i) *meta-requirements*, (ii) *designed mechanisms* in relation to the HI interaction, and (iii) *case characteristics* in terms of moderating effects related to the nature of the vision-based task (e.g., task criticality) (cf. Section 5). All researchers were asked to iterate another cycle of individual reflection and formulate testable propositions to reflect on how these three components are related to each other as observed within their cases. In this respect, the identified mechanisms had to be translated into corresponding design principles by abstracting from particular features implemented in the CV projects (Gregor et al. 2020).

In the **sixth session**, the design principles and propositions were evaluated and harmonized. As a result, an overview of the generated multi-level design knowledge was derived spanning from high-level mechanisms and meta-requirements to case-specific features and concrete requirements.

In the **seventh session**, the results were revalidated and minor adjustments on all partial results (i.e., meta-requirements, mechanisms, design principles, design features, moderators, and testable propositions) were implemented. The required steps for future work were discussed, precisely the setup of corresponding evaluation studies towards the development of a nascent design theory. At this point, the series of workshops was paused to prepare the results for presentation and obtain external feedback from a broader community.

## 4   Overview of Regarded Cases

For the selection of suitable CV development projects, the principles of Yin (2017) were followed to select cases that shared common properties (e.g., implementation of AI technology) while differing from each other to obtain the required variance (e.g., vision-based task, application domain). This non-probability sampling technique is similar to comparison focused sampling in which cases are selected to compare, contrast, and learn about characteristics that explain their similarities and differences (Saunders et al. 2009). On this basis, a total of six different cases were selected, which we describe in the following. The reported characteristics were discussed (cf. Figure 1) until the experts agreed that the characteristics both represent the individual cases appropriately and can be compared across cases (Table 2). All cases were conducted in cooperation with industry partners. Some projects are completed, while others are still under continuous development.

**Car configuration (CAR).** Car manufacturers offer their customers car configurators to assess different combinations of characteristics like color, rims, and headlights. As the 3D car rendering process is highly complex, the rendering software can output virtual car models with black holes instead of the chosen part. Currently, these faulty virtual car models are identified manually. This is highly inefficient as there are billions of possible combinations. A CV-HIS was built to detect faulty virtual car models. The CV-HIS relies on active learning to reduce the labeling effort.

**Energy infrastructure (NRG).** At present, power line maintenance relies mainly on human inspection via manual ground visitation, helicopter-based patrolling, and tower climbing. This is costly, time consuming, and often hazardous. A CV-HIS was developed that detects faults like bird nests or open safety pins on image data acquired by unmanned aerial vehicles. The detected faults are presented to human operators as a basis for decisions like prioritization and route optimization.

**Solar panels (SOL).** Manufacturers of solar panels must meet high quality standards when offering their products on the market. It is therefore important that quality impairments are detected at an early stage in the manufacturing process to treat them accordingly and avoid unnecessary costs. Thus, a CV-HIS was developed in which defects had to be detected automatically based on electroluminescence images. The challenge here was to separate defective solar cells from flawless ones and distinguish between specific types of defects while ensuring low inference times as determined by the rigid setting of the production environment.

**Viticulture (VIT).** Apart from planting crops, harvesting grapes, and producing vine, viticulture is faced with many tasks and obstacles, such as control of perfect planting positions, disease detection, and personnel allocation, especially in the harvesting process. To support these tasks with a low-cost structure, a CV-HIS was deployed that was integrated into the daily processes with minimal additional effort by mounting cameras on farm tractors to capture image data of vines and grapes to be subsequently used for disease detection and yield prognosis.

**Cutting tools (CUT).** In machining processes, unwanted material is removed from a workpiece by a cutting tool. Different types of wear occur on the tools due to friction and heat, so over time the tools are rendered unusable. A frequent task in the machining industry is the visual inspection of cutting tools. It serves as a decision basis for developing new generations of tools as well as for optimizing the parameters of machining processes. For the decisions, it is important to know precisely where which type of wear occurred. Here, a CV-HIS was developed that performs this visual analysis.

**Architectural floor plans (ARC).** The architecture, engineering, and construction industry often relies on floor plans only available as rasterized images or analog documents. For tasks such as pricing services and building operations, information such as symbols, room size, or space use must be extracted. A CV-HIS was developed to automate the digitization and analysis of floor plans. Domain experts were included to manage unknown symbols and uncertain predictions.

|  | **CAR** | **NRG** | **SOL** | **VIT** | **CUT** | **ARC** |
|---|---|---|---|---|---|---|
| **Domain** | Automotive | Energy | Manufacturing | Agriculture | Manufacturing | Architecture |
| **Objective** | Quality assurance | Fault detection | Quality assurance | Yield prediction and improvement | Understanding behavior in use of cutting tools | Digitizing and analyzing floor plans |
| **Vision-based task** | Detection | Detection | Classification, detection, segmentation | Detection, counting, object tracking | Segmentation | Detection, counting |
| **CV experience of user(s)** | Low | Medium | Medium | Low | Medium | Low |
| **Task complexity** | Medium | High | Low - medium | Low | Medium | Medium |
| **Task criticality** | Low | High | Medium | Low | Medium | Medium |
| **Exemplary requirement** | Reduction of highly repetitive manual work | Visualization of automatically extracted information | High accuracy and very low inference time | Model decisions modifiable by humans | Access to images and CV output as input for human decision-making | Flexibility regarding objects to be detected |
| **Exemplary design feature** | End-to-end automation | Interface for visual data exploration | Repository of models for different defect types | Object tracking for counting in image sequences | Human-centric decision support | Monte Carlo dropout as basis for uncertainty measures |
| **References** | Hemmer et al. (2021) | Landwehr et al. (2021) | Zschech et al. (2020) | Heinrich, Zschech, et al. (2019) | Treiss et al. (2020); Walk et al. (2020) | Hemmer, Vössing, et al. (2021) |

*Table 2.       Overview of cases.*

## 5 Conceptualization of CV-based Hybrid Intelligence Systems

To aid the design process for CV-HIS, we integrated HI theory with findings from the cases to depict a conceptualization that forms the basis of our design knowledge. The CV-HIS is conceptualized as an interaction of the *human* and the *computer* to solve a vision-based task in alignment with granular case requirements. The conceptual model is depicted in Figure 2 distinguishing between three components.

First, case requirements could be observed and linked to generic theoretical constructs to obtain *meta-requirements* determining the need, and likewise, the usefulness of any designed system functionalities.

Second, it was possible to observe several mechanisms within the CV-HIS interaction patterns between humans and computers as introduced by specifically implemented design choices, which could also be justified by mechanisms of related kernel theories. We can relate these mechanisms to three aspects that act as key resources in AI-based systems: *data*, *model*, and *decision* (Thiebes et al. 2020). The computer provides an *automation mechanism* that uses labeled image data to learn a model that generates a decision in the form of recognized objects (e.g., through the presentation of bounding box coordinates). Since the computer employs black-box models like deep neural networks, there exists a natural information asymmetry between the human and the computer. The human cannot fully comprehend the inner decision logic (i.e., the model) of the computer, and thus the computer is required to provide transparency and reduce information asymmetry by using a *signaling mechanism* (e.g., visualizing the relation between data and decisions). Additionally, the human wants to maintain some control within the decision process through a *modification mechanism,* which, for example, allows to modify data labels or manually change the computer's decision. Following the idea of a beneficial symbiosis in HI systems, the proposed mechanisms need to be coordinated by a *collaboration mechanism* that provides an interaction design via push and pull principles so that both sides can request resources (e.g., the computer can employ an active learning approach and signal the need for additional labels as a data modification from the human).

Third, several *case characteristics*, particularly related to the nature of the vision-based task, could be identified as potential moderating factors that are likely to have an effect on the relevance and need for any designed system functionality. This includes characteristics like the task's criticality and complexity or the CV experience of a user, as exemplarily summarized in Table 2.

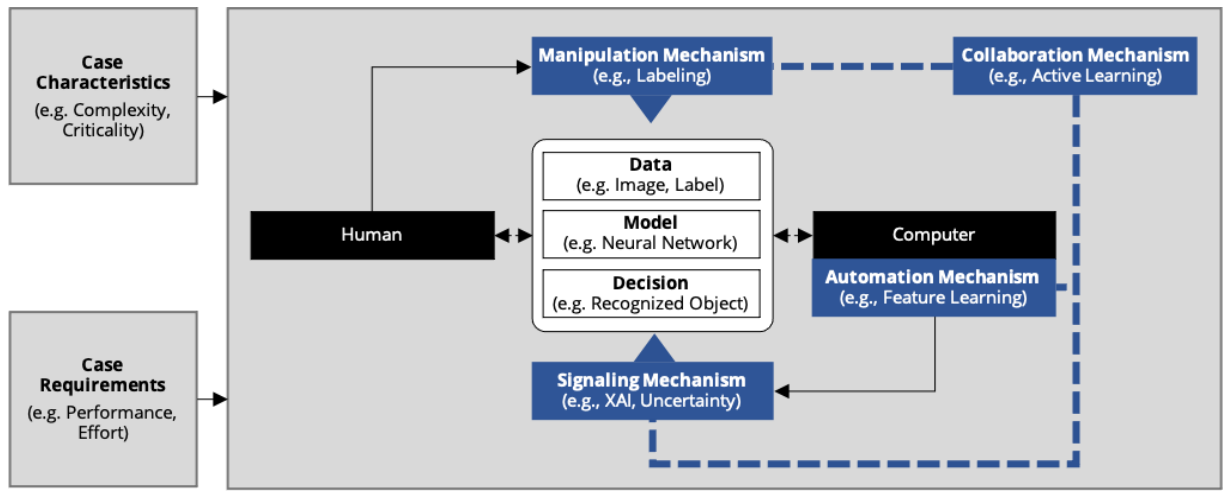

*Figure 2.*     *Conceptualization of CV-based hybrid intelligence systems.*

## 6 Design Knowledge of CV-based Hybrid Intelligence Systems

In this section, we present our results on the accumulated design knowledge of CV-HIS. Due to space limitations, we focus on the presentation of (i) meta-requirements and their connection to case requirements and related kernel theories, as well as (ii) mechanisms introduced by design in conjunction with design principles and exemplary features as implemented in the cases. A summary of the derived elements and their relationships is depicted in Figure 3. In the following, we provide further details by organizing our findings in relation to the identified mechanisms. Moreover, we present some exemplary propositions and the role of observed moderators at the end of this section.

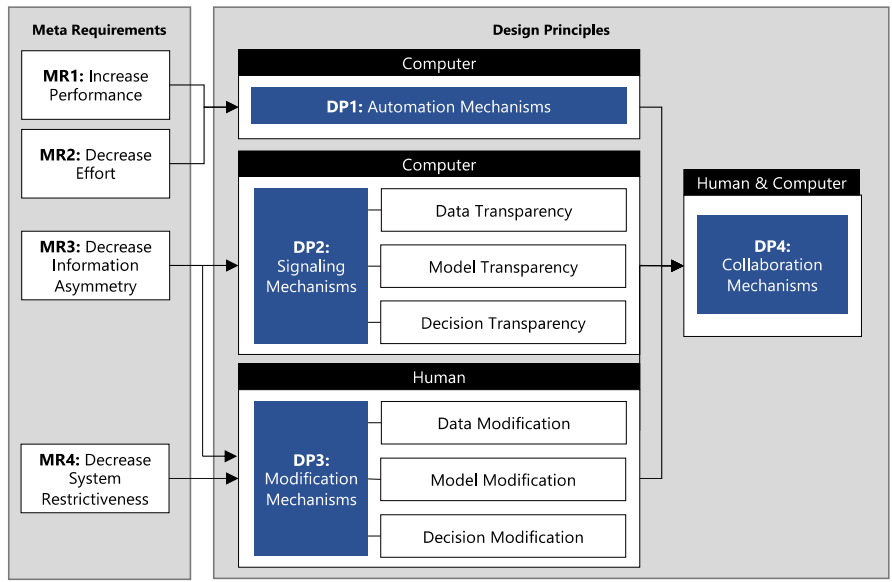

*Figure 3.        Meta-requirements and design principles for CV-based hybrid intelligence systems.*

## 6.1    Automation Mechanism

When examining the extracted requirements from the individual cases, it became apparent that the main requirement of AI-based CV systems is to reduce the manual effort to perform the vision-based tasks. Depending on the baseline situation, *effort* can take on different dimensions. In the VIT case, for example, there was no system support before so that previously vineyard objects had to be inspected and counted manually. In other cases, effort refers to the operations and configurations during system usage to perform the vision-based task. In the SOL case, for example, electroluminescence images were previously inspected by domain experts via techniques that required costly image engineering efforts that had to be reduced.

Another important requirement across all cases concerned the *performance* of the system. As such, it was necessary to automatically process and analyze visual objects while achieving a certain degree of quality, which was often required to be close or even better than human performance. This requirement was expressed by several dimensions, such as model accuracy, inference time, and required resources. In the ARC case, for example, almost 100% detection accuracy was required for correct order generation and pricing. In the SOL case, a defect detection accuracy of 98% was demanded while simultaneously guaranteeing inference times less than two seconds. In the CUT case, on the other hand, the users did not press for perfect predictions as long as they were good enough and cheaper than human experts.

Effort expectancy and performance expectancy are also pivotal factors within the unified theory of acceptance and use of technology (UTAUT). Among several other determinants, such as social influence and facilitating conditions, both factors play a significant role whether users adopt and use a new system (Venkatesh et al. 2012). Since this effect can also be assumed in the given context, we can derive the following two meta-requirements (MR) in-line with theory and observations from practice:

**MR1:**   *The vision-based task should be supported by system functionalities that improve the overall task performance (e.g., detection accuracy).*

**MR2:**   *The vision-based task should be supported by system functionalities that decrease the human effort required to perform the task (e.g., manual inspections).*

The mechanism, introduced by design, which essentially addresses these two requirements and thus constitutes the main strength of the computer within the HI interaction is the *automation mechanism*. This is about automatically extracting and processing useful information from image data that can be exploited for the respective vision-based task. As concrete system features, implementing this mechanism, different types of DL models could be identified across the six cases that are basically all based on convolutional neural network (CNN) architectures. The nested design of CNNs allows them to be fed with high-dimensional raw data and then automatically discover internal representations at different levels of abstraction that are needed for visual recognition tasks (LeCun et al. 2015). On this basis, CV tasks can be executed with high quality results while outperforming conventional types of CV systems, such as statistical approaches or shallow ML. Moreover, in contrast to conventional systems, there is no need for extensive data preparation, especially with regard to manual feature engineering, thus minimizing undesired human effort. The only central prerequisite for DL models is the availability of sufficiently large training data with labeled instances so that they can automatically recognize relevant structures. In summary, we therefore formulate the following design principle (DP) addressing the automation mechanism.

**DP1:** *Provide the system with the functionality to automatically extract visual features from image data and build a model that supports the vision-based task to minimize undesirable manual interventions.*

This design principle constitutes a core principle of today's CV systems, as it can also be observed in broader practice. It has a remarkable influence on all other design-related components, such as (meta) requirements, design features and their abstraction towards design principles. This includes, for example, that DL models generally show black-box characteristics limiting their interpretability, or that they are prone to biases induced into training data by undesired effects, which demand for further mechanisms to address such issues (Janiesch et al. 2021).

## 6.2 Signaling Mechanisms

Besides the main requirements of automation, the individual case requirements revealed additional needs concerning the reduction of information asymmetry between the human and the computer within the CV-HIS. In the VIT case, for example, a comparison between the detection of vines and grapes by the AI and the actual input data and bounding boxes was required to rely on the yield prognosis. Similarly, system users in the SOL case asked for explanations on which basis automatically detected errors were classified in one class or another. Other examples could be found in the SOL, NRG, and CUT cases, where users asked for input data visualizations to inspect training samples and labels to reduce uncertainty with regard to the actual labeling process.

Adopting principal agency theory, we can state that there is an information asymmetry between the human and the computer in CV-HIS that needs to be resolved or reduced (Vladeck 2014). On the one hand, the human has the meta-knowledge of what constitutes a real-world object that should be detected from image data by the computer, whereas the computer itself does not have such knowledge due to the lack of superintelligence (Jebari and Lundborg 2020). On the other hand, the trained system outputs predictions that are not comprehensible for the human due to the black-box-nature of the automation mechanism (Wanner et al. 2020). Failure to reduce this information asymmetry can result in decreased system adoption (Castelo et al. 2019; Miller 2019; Oh et al. 2017). Thus, connecting the theory with the case requirements, we can derive the following meta-requirement:

**MR3:** *The vision-based task should be supported by system functionalities that reduce information asymmetry between the human and the computer.*

Thus, to comply with the requirement, signaling mechanisms can be introduced by design that reduce uncertainty by creating explanations for the different aspects of the CV-HIS (i.e., data, model, and decision). The first set of explanations are aimed at increasing *data transparency*. In CV problems, the user is usually faced with detecting objects and classifying them into several available classes. The user needs to be able to determine if the input data along with its labeling is in alignment with his or her understanding of the problem domain to provide a correct input to the computer to be processed (e.g., only providing image counts as labels for criminal surveillance where specific persons are sought after will not suffice (Barbosa and Chen 2019)). Specific features related to this aspect are visualizations of input data together with suggested labels via labeling tools or providing metadata on input images. The second set of explanations covers *model transparency*. This may include to provide global explanations and meta-information regarding the trained model (e.g., configurations, hyperparameters) that are useful for a user to better comprehend the system's behavior. Lastly, local explanations are required to create *decision transparency*. The human should be able to compare the provided explanation with his or her decision logic to assess the quality of the system's decision. Methods from the field of explainable AI like LRP (layer-wise relevance propagation) or Grad-CAM (gradient-weighted class activation mapping) can be implemented for this purpose to generate pixel heat maps highlighting important parts of the input image that are responsible for the prediction. Enhancing the computer with the ability to express its confidence in single predictions is another option to reduce decision transparency by implementing features such as uncertainty measures based on, e.g., Monte Carlo dropout (Gal and Ghahramani 2016). In summary, we can thus formulate the following design principle:

**DP2:** *Provide the system with the functionality to generate and signal explanations about the computer's behavior in terms of data, model, and/or decision to increase transparency for the human.*

Figure 4 shows exemplary images, labels, predictions, and uncertainty maps from the CUT case. In the uncertainty maps, a higher degree of uncertainty is indicated by brighter pixels. The left side depicts a situation with high prediction quality. In the uncertainty map, only the pixels at the borders between classes are bright/uncertain - classifying these pixels is also difficult for humans (Kendall et al. 2015). The images on the right, on the other hand, depict a situation with low prediction quality where whole areas are bright. Consequently, the system indicates that it is uncertain about the situation, and thus, reaches its limits for correctly inspecting the cutting tool.

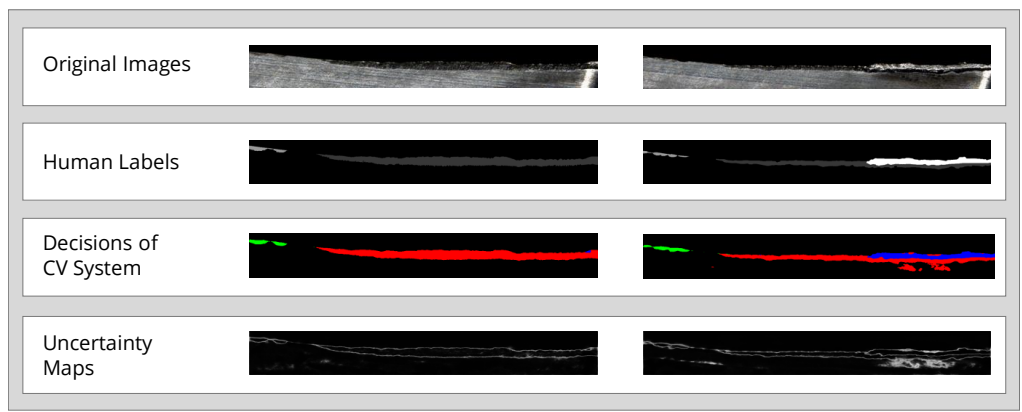

*Figure 4.     Uncertainty-based signaling in the CUT case (Treiss et al. 2020).*

## 6.3 Modification Mechanisms

On top of the previously discussed mechanisms, we could observe that the human using the developed CV-HIS did not only ask for more transparency to better understand the systems' behavior but that they also wanted to be in control of the situation they were facing. For example, in the CUT and ARC cases, professionals asked for the explicit possibility to intervene and make decisions based on their own experience when uncertain outcomes are indicated by the system. Furthermore, in the SOL and VIT cases, we could observe the requirement to have the option to directly adjust faulty labels when inspecting input data in order to achieve better model quality. Another example is given in the SOL case where the users asked to have control over the selection and configuration of detection models to choose a suitable approach depending on the situation (e.g., available resources). With regard to informing theories, our observations can be related to two theoretical lenses; that is algorithm aversion and decision support system (DSS) theory. Algorithm aversion describes that individuals are hesitant to rely on predictions computed by algorithms and rather prefer human forecasters because humans tend to believe they have superior reasoning in comparison to algorithms. However, this effect is not always present. Instead, algorithm aversion can be reduced by giving humans (even a slight degree) of control over the algorithm to modify its prediction (Dietvorst et al. 2018). Similarly, DSS theory states that a system should not be too restrictive when preparing decisions in order to not negatively limit the users' decision strategy and thus allow sufficient control over the situation (Meth et al. 2015). To sum up, we can therefore derive the following meta-requirement:

**MR4:** *The vision-based task should be supported by system functionalities with minimal restrictiveness facilitating a sufficient degree of flexibility for the human user.*

To address this requirement, it is therefore necessary to give the user sufficient flexibility and control over the system by providing appropriate *modification mechanisms.* In analogy to the manifestation of the signaling mechanisms, modification possibilities can also be designed at multiple touchpoints between the human and the computer (i.e., data, model, and decision components). Considering the *modification of data*, e.g., the designers of the SOL case developed an integrated labeling tool that was closely connected to the actual detection system. In this way, domain experts could quickly enter the image repository at any time to refine bounding boxes or modify class affiliations whenever it was necessary according to their expert knowledge. Referring to another example introduced above, it was a crucial design element within the CUT and ARC cases to integrate human knowledge in uncertain situations, which is related to the *modification of the decision*. As such, we can formulate the following design principle:

**DP3:** *Provide the system with the functionality to modify the computer's behavior in terms of data, model, and/or decision to allow the human to contribute knowledge and to control the vision-based task.*

### 6.4 Collaboration Mechanism

As outlined by Dellermann, Ebel, et al. (2019), collaboration between human users and the computer is an important characteristic of HI. It was found from the AUT and ARC cases that a collaboration mechanism can be crucial for the overall task performance. The collaboration mechanism raises awareness about a collaborator's activity and subsequently enables input requests with regard to the different mechanisms and aspects (e.g., the human could request an explanation for a specific decision from the computer). Hence, we formulate the following design principle:

**DP4:** *Provide the system with the functionality to facilitate a collaboration of human and computer-based mechanisms.*

The collaboration mechanism enables the interplay of the CV-HIS mechanisms by providing collaboration functionalities such as push and pull requests. For example, when individual decisions are uncertain (DP2 - decision transparency), the CV system can request the user to modify the provided decision where appropriate (DP3 - decision modification). This is an important aspect of the CV-HIS developed in the ARC case. The user can also support the learning process of the computer by utilizing his or her knowledge to label unknown data (DP3 - data modification) where the system expects the highest level of improvement (DP2 - model transparency). This pattern, frequently referred to as *active learning,* is a well-known example of dialogue-based collaboration mechanism and is a crucial design element within the AUT case. Another example of such a combination is given in the SOL case, where users can also choose between different levels of model complexity (DP3 - model modification) based on the task characteristics and the computer's signaling outcome (DP2 - model transparency).

### 6.5 Outlook: Exemplary Propositions and Moderating Effects

With the accumulated design knowledge outlined in the previous sections, it is possible to derive a set of testable propositions (TP) constituting the relationships between meta-requirements and design principles. Since we cannot fully discuss all derived propositions due to space limitations, we only provide two examples by TP1 and TP2 to pave the further way towards a nascent design theory.

**Example Proposition 1:** *Using a CV-HIS with automation mechanisms (DP1) will result in a higher degree of perceived performance than using a system without automation mechanisms.*

**Example Proposition 2:** *Using a CV-HIS with automation mechanisms (DP1) and modification mechanisms (DP3) will result in a higher degree of perceived control than using a CV-HIS with automation mechanisms, but without modification mechanisms.*

Both propositions reflect generalized design knowledge as observed within the cases and place it in relation to a respective reference system. Thus, TP1 describes the effect of employing a system with the core design principle DP1 in contrast to another type of CV system (e.g., conventional system based on statistical approaches), whereas TP2 describes the effect of two different design configurations with regard to DP1 and DP3. Thus, with the different meta-requirements and design principles, a system of propositions can be obtained that describes the effects of the mechanisms introduced by design in their entirety. However, since not all mechanisms could be observed equally across all cases, the propositions are only tentative assumptions that need to be examined in larger evaluation studies in more controllable settings and with more users involved.

Furthermore, we assume that the contextual case characteristics, such as the criticality of the vision-based task or the CV experience of users, might have a significant influence on the intensity of the need for any design principle. Thus, we assume, for example, that there is a higher need for explanation (DP2) and modification (DP3) of generated decisions in cases where domain professionals are responsible for critical situations that can lead to high costs or even lives at risk. Similarly, we expect that CV cases with rather technically oriented users (e.g., SOL case) will presumably require a higher degree of control over models and configurations (DP3) so that they can bring their technical expertise into the processes than it will be the case, for example, in agricultural domains (e.g., VIT case). While some of these characteristics and their influences were already captured in this research project, such moderating effects need to be further examined in future studies.

# 7 Conclusion

In this paper, we provided insights from a research project with the goal to accumulate prescriptive design knowledge for computer-vision-based hybrid intelligence systems. To this end, we pursued a reflective DSR approach, introduced as "Strategy II" by Iivari (2015). We conducted a series of workshops with IS researchers involved in several industrial computer vision projects. As a result, we were able to derive generalizable design knowledge illustrated through meta-requirements, mechanisms, design principles as well as testable propositions. Even though our focus was on the application area of computer vision, we are confident that the results show a more generalizable character and can therefore be transferred to broader contexts in which hybrid intelligence systems need to be designed (e.g., in the realm of natural language processing).

However, the generalizability of these results is subject to certain limitations. For instance, we only regarded a total of six cases and future work should include additional examples to further facilitate the theorizing process. Furthermore, due to space restrictions, we could only elaborate on some excerpts of the current findings. For instance, we had to exclude design principles with a more technical focus (like scalability and robustness), and we could not discuss the potential moderating effects and testable propositions in sufficient detail, which will therefore be part of subsequent work.

Our work can help practitioners to design CV-HIS in a more human-centric manner by incorporating socio-technical considerations. From an academic point of view, this research contributes to the knowledge base by proposing generalizable design knowledge and laying the foundation for many future research directions in need of further investigation by considering factors beyond technical performance like restrictiveness and information asymmetry between human and computer. Future work needs to focus on the testable propositions and their translation into evaluation studies and experiments. A promising field of research lies ahead.